\documentclass[10pt,twocolumn,letterpaper]{article}

\usepackage{cvpr}
\usepackage{times}
\usepackage{epsfig}
\usepackage{graphicx}
\usepackage{amsmath}
\usepackage{amssymb}
\usepackage{multirow}
\usepackage{threeparttable}
\usepackage{bbding}
\usepackage{booktabs} 
\usepackage{url}
\def\eg{{\em e.g.}}
\def\ie{{\em i.e.}}
\usepackage[pagebackref=true,breaklinks=true,letterpaper=true,colorlinks,bookmarks=false]{hyperref}

\cvprfinalcopy 
\def\cvprPaperID{503} 

\ifcvprfinal\fi
\begin{document}

\title{A Dataset and Benchmark for Large-scale Multi-modal Face Anti-spoofing}
\author{Shifeng Zhang$^{\rm 1*}$, Xiaobo Wang$^{\rm 2}$\thanks{These authors contributed equally to this work},\ \ Ajian Liu$^{\rm 3}$, Chenxu Zhao$^{\rm 2}$, \\Jun Wan$^{\rm 1}$\thanks{Corresponding author},\ \ Sergio Escalera$^{\rm 4}$, Hailin Shi$^{\rm 2}$, Zezheng Wang$^{\rm 5}$, Stan Z. Li$^{\rm 1,3}$\vspace{1.2mm}\\
{$^{\rm 1}$NLPR, CASIA, UCAS, China;
$^{\rm 2}$JD AI Research;
$^{\rm 3}$MUST, Macau, China}\\
{$^{\rm 4}$Universitat de Barcelona, Computer Vision Center, Spain;
$^{\rm 5}$JD Finance}\\
{\tt\small
\{shifeng.zhang,jun.wan,szli\}@nlpr.ia.ac.cn, ajianliu92@gmail.com\vspace{-1.2mm}
}\\
{\tt\small
\{wangxiaobo8,zhaochenxu1,shihailin,wangzezheng1\}@jd.com, sergio@maia.ub.es
}\\
}
\maketitle
\thispagestyle{empty}

\begin{abstract}
Face anti-spoofing is essential to prevent face recognition systems from a security breach. Much of the progresses have been made by the availability of face anti-spoofing benchmark datasets in recent years. However, existing face anti-spoofing benchmarks have limited number of subjects ($\le\negmedspace170$) and modalities ($\leq\negmedspace2$), which hinder the further development of the academic community. To facilitate face anti-spoofing research, we introduce a large-scale multi-modal dataset, namely CASIA-SURF, which is the largest publicly available dataset for face anti-spoofing in terms of both subjects and visual modalities. Specifically, it consists of $1,000$ subjects with $21,000$ videos and each sample has $3$ modalities (\ie, RGB, Depth and IR). We also provide a measurement set, evaluation protocol and training/validation/testing subsets, developing a new benchmark for face anti-spoofing. Moreover, we present a new multi-modal fusion method as baseline, which performs feature re-weighting to select the more informative channel features while suppressing the less useful ones for each modal. Extensive experiments have been conducted on the proposed dataset to verify its significance and generalization capability. The dataset is available at \url{https://sites.google.com/qq.com/chalearnfacespoofingattackdete/}.
\end{abstract}

\section{Introduction}
Face anti-spoofing aims to determine whether the captured face of a face recognition system is real or fake. With the development of deep convolutional neural network (CNN), face recognition~\cite{Bhattacharjee_BTAS2018_2018,chi2018selective,Mohammadi2018Deeply,wang2018ensemble,zhang2019single} has achieved near-perfect recognition performance and already has been applied in our daily life, such as phone unlock, access control, face payment, etc. However, these face recognition systems are prone to be attacked in various ways, including print attack, video replay attack and 2D/3D mask attack, which cause the recognition result to become unreliable. Therefore, face presentation attack detection (PAD)~\cite{Boulkenafet2016Face,Boulkenafet2017Face} is a vital step to ensure that face recognition systems are in a safe reliable condition.

\begin{figure}[t]
\centering
\includegraphics[width=1.0\linewidth]{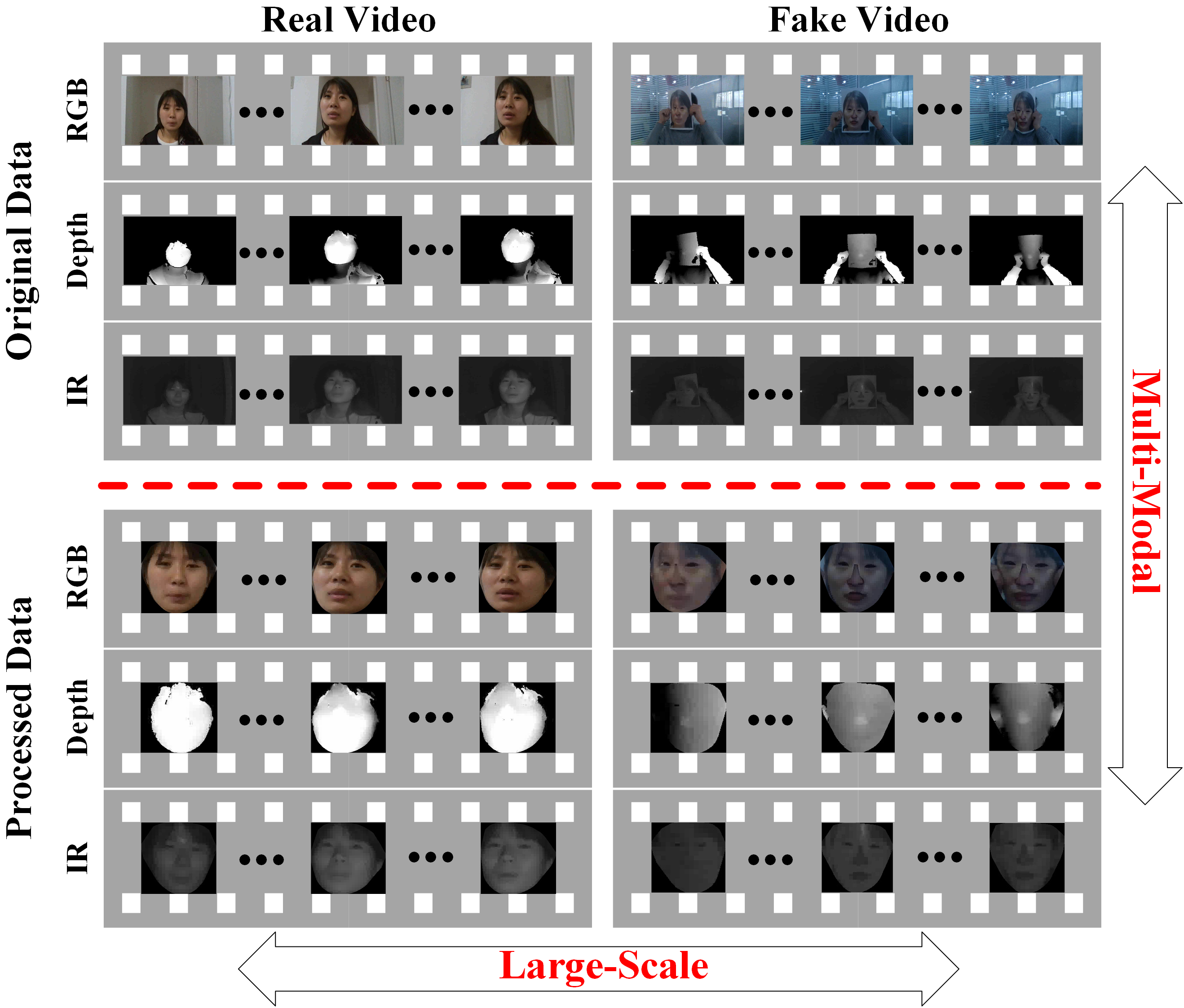}
\caption{The CASIA-SURF dataset. It is a large-scale and multi-modal dataset for face anti-spoofing, consisting of $492,522$ images with $3$ modalities (\ie, RGB, Depth and IR).}
\label{fig:dataset}
\end{figure}

\begin{table*}
\renewcommand\arraystretch{1.15}
\centering
\setlength{\tabcolsep}{13pt}
\footnotesize{
\begin{tabular}{|c|c|c|c|c|c|c|c|}
\hline
Dataset &Year &\# of subjects & \# of videos & Camera & Modal types&  Spoof attacks \\
\hline
\hline
Replay-Attack~\cite{Chingovska_BIOSIG-2012} & 2012 & 50&   1200  & VIS& RGB&  Print, 2 Replay\\
\hline
CASIA-MFSD~\cite{Zhang2012A}                & 2012 & 50&   600   & VIS& RGB&  Print, Replay\\
\hline
3DMAD~\cite{ERDOGMUS_BTAS-2013}             & 2013 & 17&   255    & VIS/Kinect&   RGB/Depth&     3D Mask\\
\hline
MSU-MFSD~\cite{Wen2015Face}                 & 2015 & 35&   440    & Phone/Laptop& RGB&  Print, 2 Replay\\
\hline
Replay-Mobile~\cite{Costa2016The}           & 2016 & 40&   1030   & VIS&  RGB&  Print, Replay\\
\hline
Msspoof~\cite{msspoof-2015}                 & 2016 & 21&   4704$^{*}$    & VIS/NIR&      RGB/IR&        Print\\
\hline
Oulu-NPU~\cite{Boulkenafet2017OULU}         & 2017 & 55&   5940 & VIS&          RGB&  2 Print, 2 Replay\\
\hline
SiW~\cite{Liu2018Learning}                  & 2018 & 165&  4620 & VIS&          RGB&  2 Print, 4 Replay\\
\hline
\textbf{CASIA-SURF (Ours)}                        & 2018 & \textbf{1000} & \textbf{21000} &  RealSense &   \textbf{RGB/Depth/IR} &  Print, Cut\\
\hline
\end{tabular}
}
\caption{The comparison of the public face anti-spoofing datasets ($*$ indicates this dataset only contains images, not video clips).}
\label{table:datasets}
\end{table*}

Recently, face PAD algorithms~\cite{Jourabloo2018Face,Liu2018Learning} have achieved great performances. One of the key points of this success is the availability of face anti-spoofing datasets~\cite{Boulkenafet2017OULU,Chingovska_BIOSIG-2012,Costa2016The,Liu2018Learning,Wen2015Face,Zhang2012A}. However, compared to the large existing image classification \cite{deng2009imagenet} and face recognition \cite{yi2014learning} datasets, face anti-spoofing datasets have less than $170$ subjects and $60,00$ video clips, see Table~\ref{table:datasets}. The limited number of subjects does not guarantee for the generalization capability required in real applications. Besides, from Table~\ref{table:datasets}, another problem is the limited number of data modalities. Most of the current datasets only have one modal (\textit{e.g.}, RGB), and the existing available multi-modal datasets~\cite{ERDOGMUS_BTAS-2013,msspoof-2015} are scarce, including no more than 21 subjects. 

In order to deal with previous drawbacks, we introduce a large-scale multi-modal face anti-spoofing dataset, namely CASIA-SURF, which consists of $1,000$ subjects and $21,000$ video clips with $3$ modalities (RGB, Depth, IR). It has $6$ types of photo attacks combined by multiple operations, \eg, cropping, bending the print paper and stand-off distance. Some samples of the dataset are shown in Figure~\ref{fig:dataset}. As shown in Table~\ref{table:datasets}, our dataset has two main advantages: (1) It is the largest one in term of number of subjects and videos; (2) The dataset is provided with three modalities (\ie, RGB, Depth and IR).

Another open issue in face anti-spoofing is how performance should be computed. Many works~\cite{Liu2018Learning,Jourabloo2018Face,Boulkenafet2017OULU,Costa2016The} adopt the attack presentation classification error rate (APCER), bona fide presentation classification error rate (BPCER) and average classification error rate (ACER) as the evaluation metric, in which APCER and BPCER are used to measure the error rate of fake or live samples, and ACER is the average of APCER and BPCER scores. However, in real applications, one may be more concerned about the false positive rate, \ie, attacker is treated as real/live one. Inspired by face recognition~\cite{liu2017sphereface,wang2018support}, the receiver operating characteristic (ROC) curve is introduced for large-scale face anti-spoofing in our dataset, which can be used to select a suitable trade off threshold between false positive rate (FPR) and true positive rate (TPR) according to the requirements of a given real application.

To sum up, the contributions of this paper are three-fold: (1) We present a large-scale multi-modal dataset for face anti-spoofing. It contains $1,000$ subjects, being at least $6$ times larger than existing datasets, with three modalities. (2) We introduce a new multi-modal fusion method to effectively merge the involved three modalities, which performs modal-dependent feature re-weighting to select the more informative channel features while suppressing the less useful ones for each modality. (3) We conduct extensive experiments on the proposed CASIA-SURF dataset.

\section{Related work}
\subsection{Datasets}
Most of existing face anti-spoofing datasets only contain the RGB modalitiy. Replay-Attack~\cite{Chingovska_BIOSIG-2012} and CASIA-FASD~\cite{Zhang2012A} are two widely used PAD datasets. Even the recently released SiW~\cite{Liu2018Learning} dataset, collected with high resolution image quality, only contains RGB data. With the widespread application of face recognition in mobile phones, there are also some RGB datasets recorded by replaying face video with smartphone, such as MSU-MFSD~\cite{Wen2015Face}, Replay-Mobile~\cite{Costa2016The} and OULU-NPU~\cite{Boulkenafet2017OULU}.

As attack techniques are constantly upgraded, some new types of presentation attacks (PAs) have emerged, \eg, 3D~\cite{ERDOGMUS_BTAS-2013} and silicone masks~\cite{Bhattacharjee_BTAS2018_2018}. These are more realistic than traditional 2D attacks. Therefore, the drawbacks of visible cameras are revealed when facing these realistic face masks. Fortunately, some new sensors have been introduced to provide more possibilities for face PAD methods, such as depth cameras, muti-spectral cameras and infrared light cameras. Kim \textit{et al.}~\cite{Kim2009Masked} aim to distinguish between the facial skin and mask materials by exploiting their reflectance. Kose \textit{et al.}~\cite{Kose2013Countermeasure} propose a 2D+3D face mask attacks dataset to study the effects of mask attacks. However, associated data has not been made public. 3DMAD~\cite{ERDOGMUS_BTAS-2013} is the first publicly available 3D masks dataset, which is recorded using Microsoft Kinect sensor and consists of Depth and RGB modalities. Another multi-modal face PAD dataset is Msspoof~\cite{msspoof-2015}, containing visible (VIS) and near-infrared (NIR) images of real accesses and printed spoofing attacks with $\leq21$ objects.

However, existing datasets in the face PAD community have two common limitations. First, they all have the limited number of subjects and samples, resulting in a potential over-fitting risk when face PAD algorithms are tested on these datasets~\cite{Chingovska_BIOSIG-2012,Zhang2012A}. Second, most of existing datasets are captured by visible camera that only includes the RGB modality, causing a substantial portion of 2D PAD methods to fail when facing new types of PAs (3D and custom-made silicone masks).

\subsection{Methods}
Face anti-spoofing has been studied for decades. Some previous works~\cite{Pan2007Eyeblink,wang2009face,kollreider2008verifying,Bharadwaj2013Computationally} attempt to detect the evidence of liveness (\ie, eye-blinking). Another works are based on contextual~\cite{Pan2011Monocular,Komulainen2014Context} and moving ~\cite{Wang2013Face,De2012Moving,Kim2013Face} information. To improve the robustness to illumination variation, some algorithms adopt HSV and YCbCr color spaces~\cite{Boulkenafet2016Face,Boulkenafet2017Face}, as well as Fourier spectrum~\cite{Li2004Live}. All of these methods use handcrafted features, such as LBP~\cite{ojala2002multiresolution,chingovska2012effectiveness,Yang2013Face,Maatta2012Face}, HoG~\cite{Yang2013Face,Maatta2012Face,schwartz2011face} and GLCM~\cite{schwartz2011face}. They are fast and achieve a relatively satisfactory performance on small public face anti-spoofing datasets.

Some fusion methods have been proposed to obtain a more general countermeasure effective against a variation of attack types. Tronci \etal~\cite{tronci2011fusion} proposed a linear fusion of frame and video analysis. Schwartz \etal~\cite{schwartz2011face} introduced feature level fusion by using Partial Least Squares (PLS) regression based on a set of low-level feature descriptors. Other works~\cite{de2013can,komulainen2013complementary} obtained an effective fusion scheme by measuring the level of independence of two anti-counterfeiting systems. However, these fusion methods focus on score or feature level, not modality level, due to the lack of multi-modal datasets. 

Recently, CNN-based methods~\cite{feng2016integration, li2016original, Patel2016Secure, yang2014learn, Liu2018Learning, Jourabloo2018Face} have been presented in the face PAD community. They treat face PAD as a binary classification problem and achieve remarkable improvements in the intra-testing. Liu \etal. \cite{Liu2018Learning} designed a network architecture to leverage two auxiliary information (Depth map and rPPG signal) as supervision. Amin \etal. \cite{Jourabloo2018Face} introduced a new perspective for solving the face anti-spoofing by inversely decomposing a spoof face into the live face and the spoof noise pattern. However, they exhibited a poor generalization ability on the cross-testing due to the over-fitting to training data. This problem remains open, although some works~\cite{li2016original,Patel2016Secure} adopted transfer learning to train a CNN model from ImageNet~\cite{deng2009imagenet}. These works show the need of a larger PAD dataset.

\begin{figure}[t]
\centering
\includegraphics[width=1\linewidth]{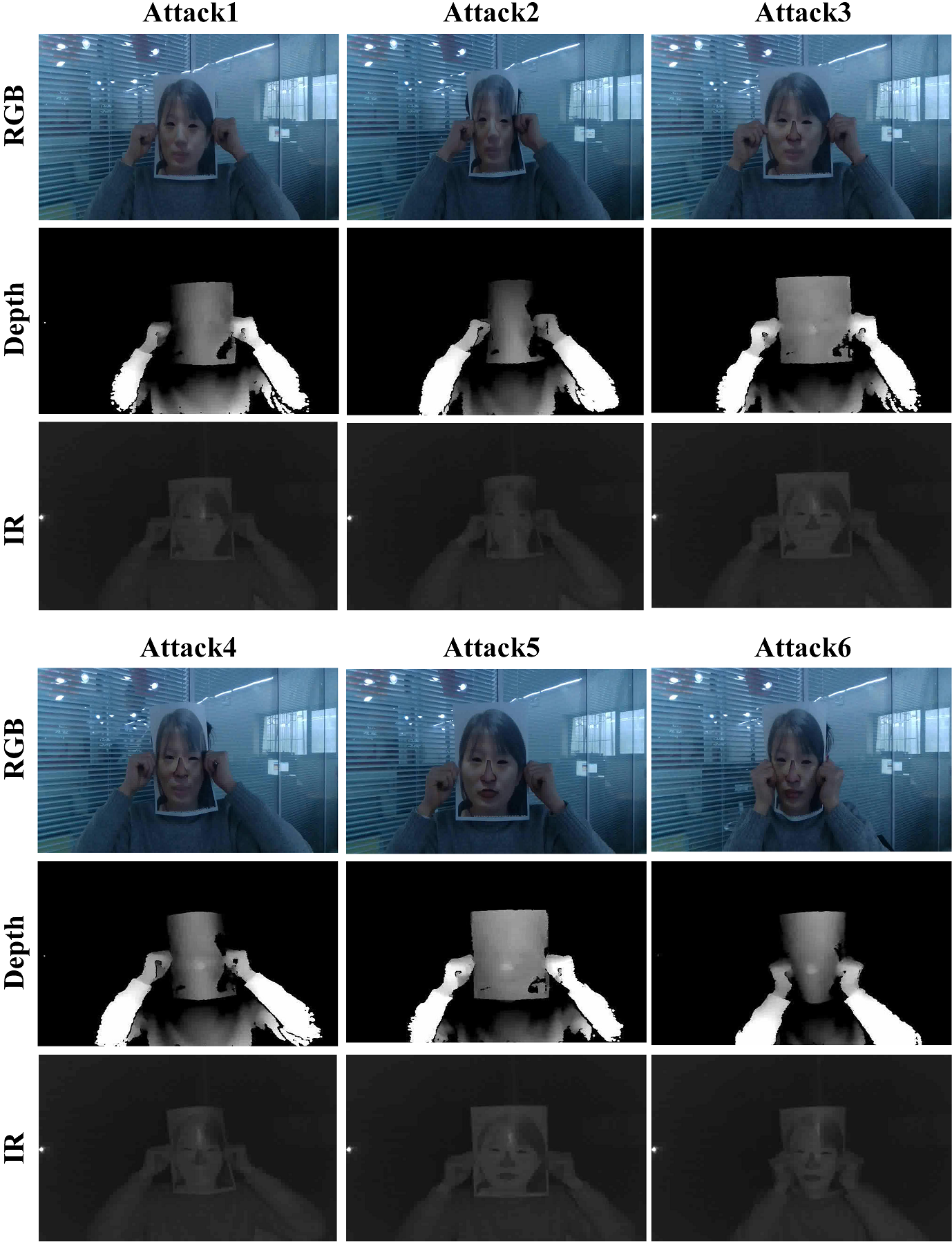}
\caption{Six attack styles in the CASIA-SURF dataset.}
\label{fig:attacks}
\end{figure}

\section{CASIA-SURF dataset}
As commented, all existing datasets involve a reduced number of subjects and just one visual modality. Although the publicly available datasets have driven the development of face PAD and continue to be valuable tools for this community, their limited size severely impede the development of face PAD with higher recognition to be applied in problems such as face payment or unlock.

In order to address current limitations in PAD, we collected a new face PAD dataset, namely the CASIA-SURF dataset. To the best our knowledge, CASIA-SURF dataset is currently the largest face anti-spoofing dataset, containing $1,000$ Chinese people in $21,000$ videos. Another motivation in creating this dataset, beyond pushing the research on face anti-spoofing, is to explore recent face spoofing detection models performance when considering a large amount of data. In the proposed dataset, each sample includes $1$ live video clip, and $6$ fake video clips under different attack ways (one attack way per fake video clip). In the different attack styles, the printed flat or curved face images will be cut eyes, nose, mouth areas, or their combinations. Finally, $6$ attacks are generated in the CASIA-SURF dataset. Fake samples are shown in Figure~\ref{fig:attacks}. Detailed information of the $6$ attacks is given below.
\begin{itemize}
\setlength{\itemsep}{5pt}
\setlength{\parsep}{0pt}
\setlength{\parskip}{0pt}
\item Attack 1: One person hold his/her flat face photo where eye regions are cut from the printed face.
\item Attack 2: One person hold his/her curved face photo where eye regions are cut from the printed face.
\item Attack 3: One person hold his/her flat face photo where eyes and nose regions are cut from the printed face.
\item Attack 4: One person hold his/her curved face photo where eyes and nose regions are cut from the printed face.
\item Attack 5: One person hold his/her flat face photo where eyes, nose and mouth regions are cut from the printed face.
\item Attack 6: One person hold his/her curved face photo where eyes, nose and mouth regions are cut from the printed face.
\end{itemize}

\begin{figure}[t]
\centering
\includegraphics[width=1\linewidth]{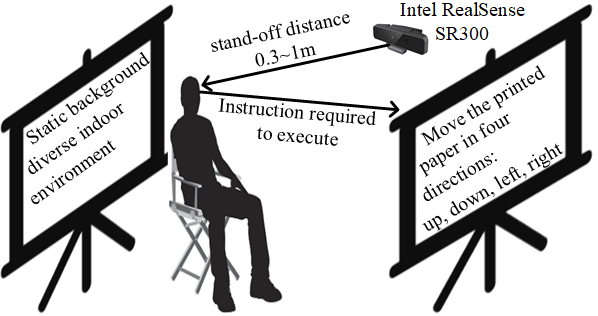}
\caption{Illustrative sketch of recordings setups in the CASIA-SURF dataset.}
\label{fig:shotting}
\end{figure}

\subsection{Acquisition details}
We used the Intel RealSense SR300 camera to capture the RGB, Depth and Infrared (IR) videos simultaneously. In order to obtain the attack faces, we printed the color pictures of the collectors with A4 paper. During the video recording, the collectors were required to do some actions, such as turn left or right, move up or down,  walk in or away from the camera. Moreover, the face angle of performers were asked to be less $30^0$. The performers stood within the range of $0.3$ to $1.0$ meter from the camera. The diagram of data acquisition procedure is shown in Figure~\ref{fig:shotting}, where it shows how the multi-modal data was recorded via Intel RealSence SR300 camera.

\begin{figure}[t]
\centering
\includegraphics[width=1\linewidth]{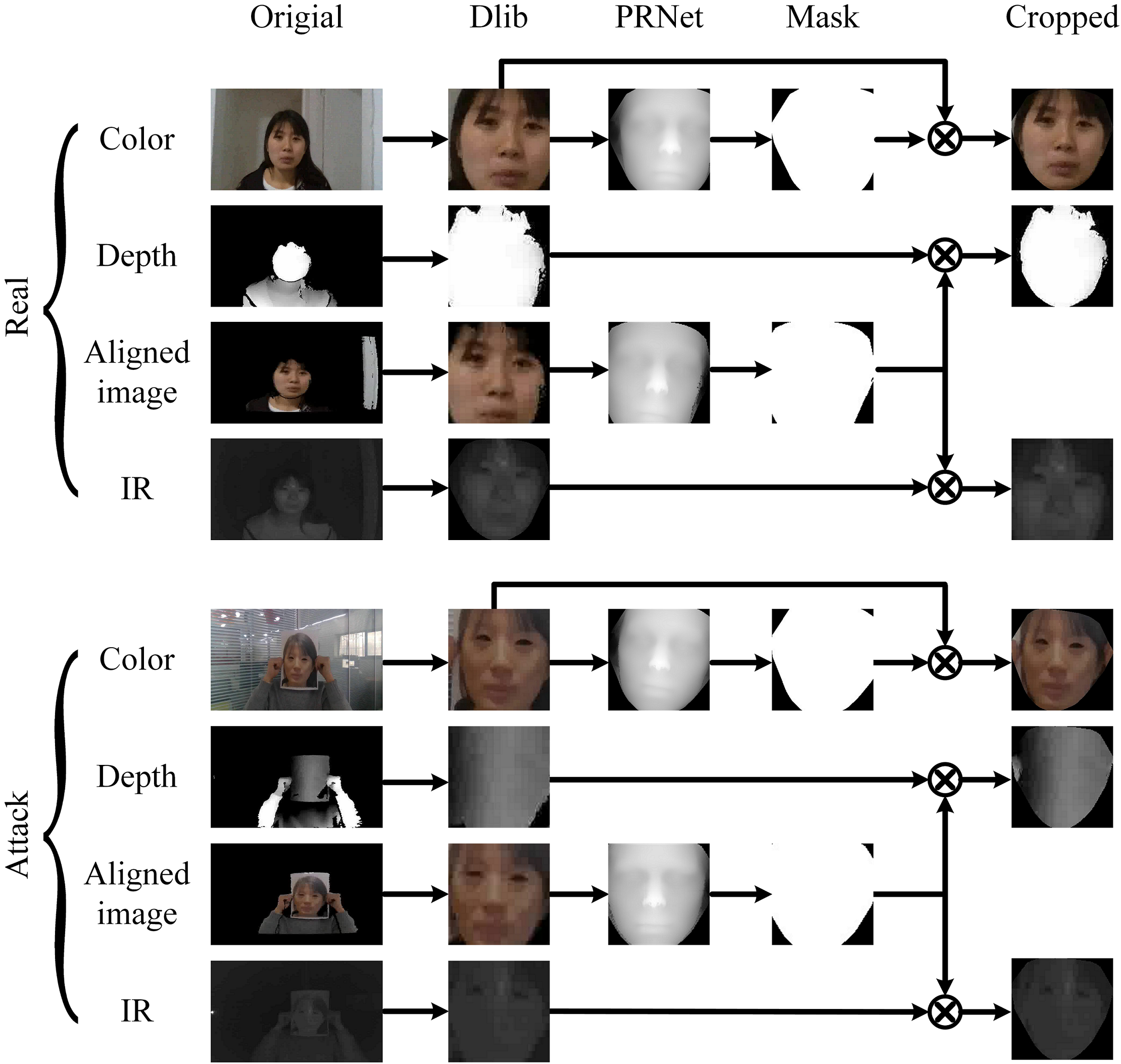}
\caption{Preprocessing details of the three modalities of the CASIA-SURF dataset.}
\label{fig:Preprocessing}
\end{figure}

Four video streams including RGB, Depth and IR images were captured at the same time, plus the RGB-Depth-IR aligned images using RealSense SDK. The RGB, Depth, IR and aligned images are shown in the first column of Figure~\ref{fig:Preprocessing}. The resolution is $1280\times720$ for RGB images, and $640\times480$ for Depth, IR and aligned images.

\subsection{Data preprocessing} \label{sec:process}
In order to create a challenging dataset, we removed the background except face areas from original videos. Concretely, as shown in Figure~\ref{fig:Preprocessing}, the accurate face area is obtained through the following steps. Given that we have a RGB-Depth-IR aligned video clip for each sample, we first used Dlib~\cite{dlib09} to detect face for every frame of RGB and RGB-Depth-IR aligned videos, respectively. The detected RGB and aligned faces are shown in the second column of Figure~\ref{fig:Preprocessing}. After face detection, we applied the PRNet \cite{DBLP:conf/eccv/FengWSWZ18} algorithm to perform 3D reconstruction and density alignment on the detected faces. The accurate face area (namely, face reconstruction area) is shown in the third column of Figure~\ref{fig:Preprocessing}. Then, we defined a binary mask based on non-active face reconstruction area from previous steps. The binary masks of RGB and RGB-Depth-IR images are shown in the fourth column of Figure~\ref{fig:Preprocessing}. Finally, we obtained face area of RGB image via pointwise product between RGB image and RGB binary mask. The Depth (or IR) area can be calculated via the pointwise product between Depth (or IR) image and RGB-Depth-IR binary mask. The face images of three modalities (RGB, Depth, IR) are shown in the last column of Figure~\ref{fig:Preprocessing}.

\subsection{Statistics}

\begin{table}[b]
\renewcommand\arraystretch{1.15}
\centering
\setlength{\tabcolsep}{4.25pt}
\small{
\begin{tabular}{|c|c|c|c|c| }
\hline
& Training & Validation & Testing & Total  \\
\hline
\# Obj.            & 300     & 100   & 600 & 1000\\
\# Videos             & 6,300   & 2,100 & 12,600 & 21000 \\
\# Ori. img.      & 1,563,919 & 501,886 & 3,109,985 & 5,175,790  \\
\# Samp. img.     & 151,635 & 49,770 & 302,559 & 503,964 \\
\# Crop. img.     & 148,089 & 48,789 & 295,644 & 492522 \\
\hline
\end{tabular}}
\vspace{0.5mm}
\caption{Statistical information of the proposed CASIA-SURF dataset.}
\label{dataset}
\end{table}

Table~\ref{dataset} presents the main statistics of the proposed CASIA-SURF dataset:

(1) There are $1,000$ subjects and each one has a live video clip and six fake video clips. Data contains variability in terms of gender, age, glasses/no glasses, and indoor environments.

(2) Data is split in three sets: training, validation and testing. The training, validation and testing sets have $300$, $100$ and $600$ subjects, respectively. Therefore, we have $6,300$ ($2,100$ per modality), $2,100$ ($700$ per modality), $12,600$ ($4,200$ per modality) videos for its corresponding set.

(3) From original videos, there are about $1.5$ million, $0.5$ million, $3.1$ million frames in total for training, validation, and testing sets, respectively. Owing to the huge amount of data, we select one frame out of every 10 frames and formed the sampled set with about $151K$, $49K$, and $302K$ for training, validation and testing sets, respectively.

(4) After data prepossessing in Sec.~\ref{sec:process} and removing non-detected face poses with extreme lighting conditions, we finally obtained about $148K$, $48K$, $295K$ frames for training, validation and testing sets on the CASIA-SURF dataset, respectively.

\begin{figure}[t]
\centering
\includegraphics[width=1\linewidth]{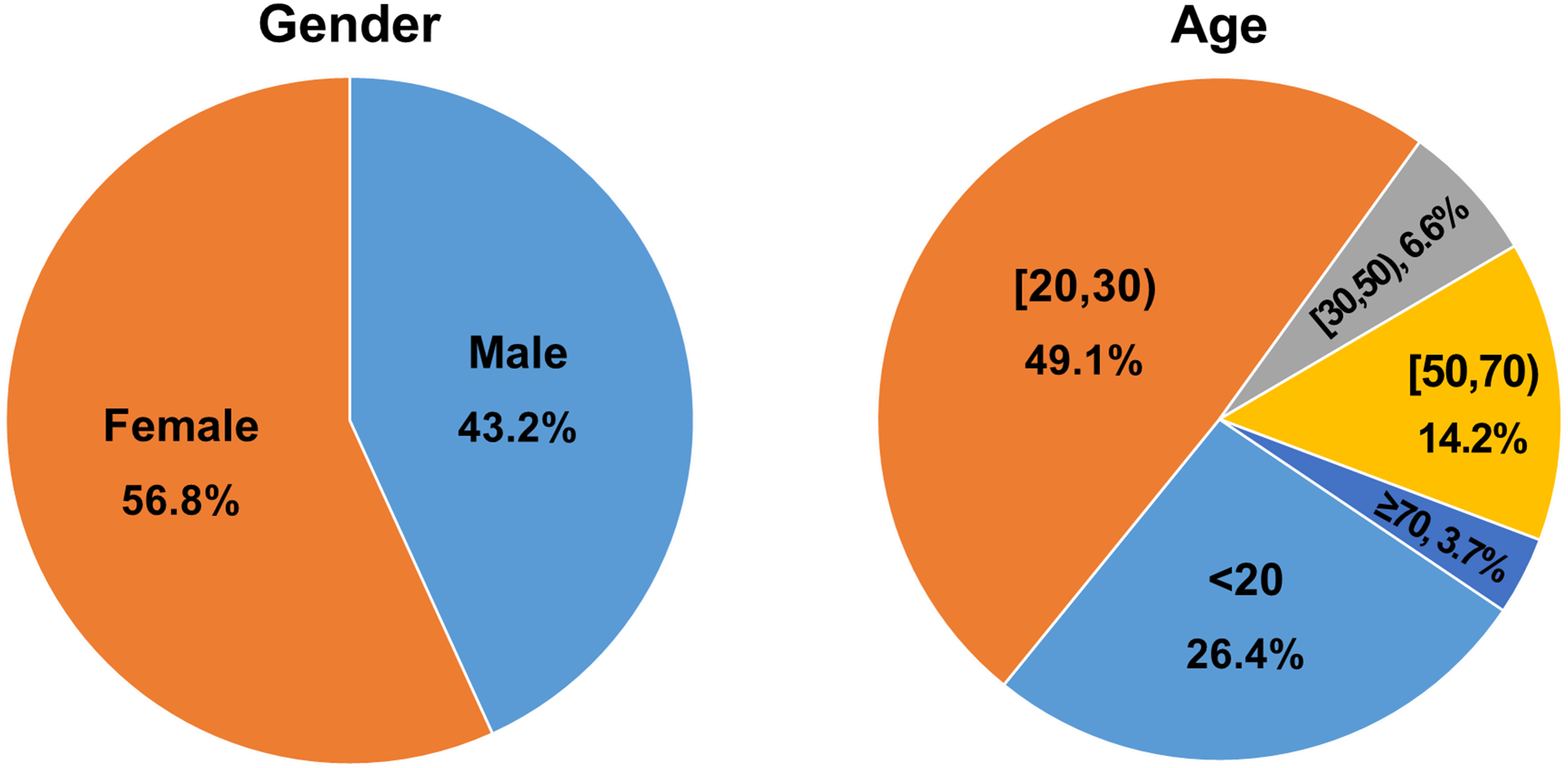}
\caption{Gender and age distribution of the CASIA-SURF dataset.}
\label{fig:statics}
\end{figure}

All subjects are Chinese, and the information of gender statistics is shown in the left side of Figure\ref{fig:statics}. It shows that the ratio of female is $56.8\%$ while the ratio of male is $43.2\%$.
In addition, we also show age distribution of the CASIA-SURF dataset in the right side of Fig \ref{fig:statics}. One can see a wide distribution of age ranges from $20$ to more than $70$ years old, while most of subjects are under $70$ years old. On average, the range of $[20,30)$ ages is dominant, being about $50\%$ of all the subjects.

\subsection{Evaluation protocol}

\noindent \textbf{Intra-testing}. For the intra-testing protocol, the live faces and Attacks 4, 5, 6 are used to train the models. Then, the live faces and Attacks 1, 2, 3 are used as the validation and testing sets. The validation set is used for model selection and the testing set for final evaluation. This protocol is used for the evaluation of face anti-spoofing methods under controlled conditions, where training and testing sets belong to the CASIA-SURF dataset. The main reason behind this selection of attack types in the training and testing sets is to increase the difficulty of the face anti-spoofing detection task. In this experiment, we show that there is still a big space to improve the performance under the ROC evaluation metric, especially, how to improve the true positive rate (TPR) at the small value of false positive rate (FPR), such as FPR=$10^{-5}$.

\noindent \textbf{Cross-testing}. The cross-testing protocol uses the training set of CASIA-SURF to train the deep models, which are then fine-tuned on the target training dataset (\textit{e.g.}, the training set of SiW \cite{Liu2018Learning}). Finally, we test the fine-tuned model on the target testing set (\textit{e.g.}, the testing set of SiW \cite{Liu2018Learning}). The cross-testing protocol aims at simulating performance in real application scenarios involving high variabilities in appearance and having a limited number of samples to train the model.

\section{Method}
Before showing some experimental analysis on the dataset, we first built a strong baseline method. We aim at finding a straightforward architecture that provides good performance in our CASIA-SURF dataset. Thus, we define the face anti-spoofing problem as a binary classification task (fake \emph{v.s} real) and conduct the experiments based on the ResNet-18 \cite{he2016deep} classification network. ResNet-18 consists of five convolutional blocks (namely res1, res2, res3, res4, res5), a global average pooling layer and a softmax layer, which is a relatively shallow network with high classification performance.

\subsection{Naive halfway fusion}
CASIA-SURF is characterized by multi-modality (\ie, RGB, Depth, IR) and a key issue is how to fuse the complementary information between the three modalities. We use a multi-stream architecture with three subnetworks to study the dataset modalities, in which RGB, Depth and IR data are learnt separately by each stream, and then shared layers are appended at a point to learn joint representations and perform cooperated decisions. The halfway fusion is one of the commonly used fusion methods, which combines the subnetworks of different modalities at a later stage, \ie, immediately after the third convolutional block (res3) via the feature map concatenation. In this way, features from different modalities can be fused to perform classification. However, direct concatenating these features cannot make full use of the characteristics between different modalities.

\begin{figure}
\centering
\includegraphics[width=0.48\textwidth]{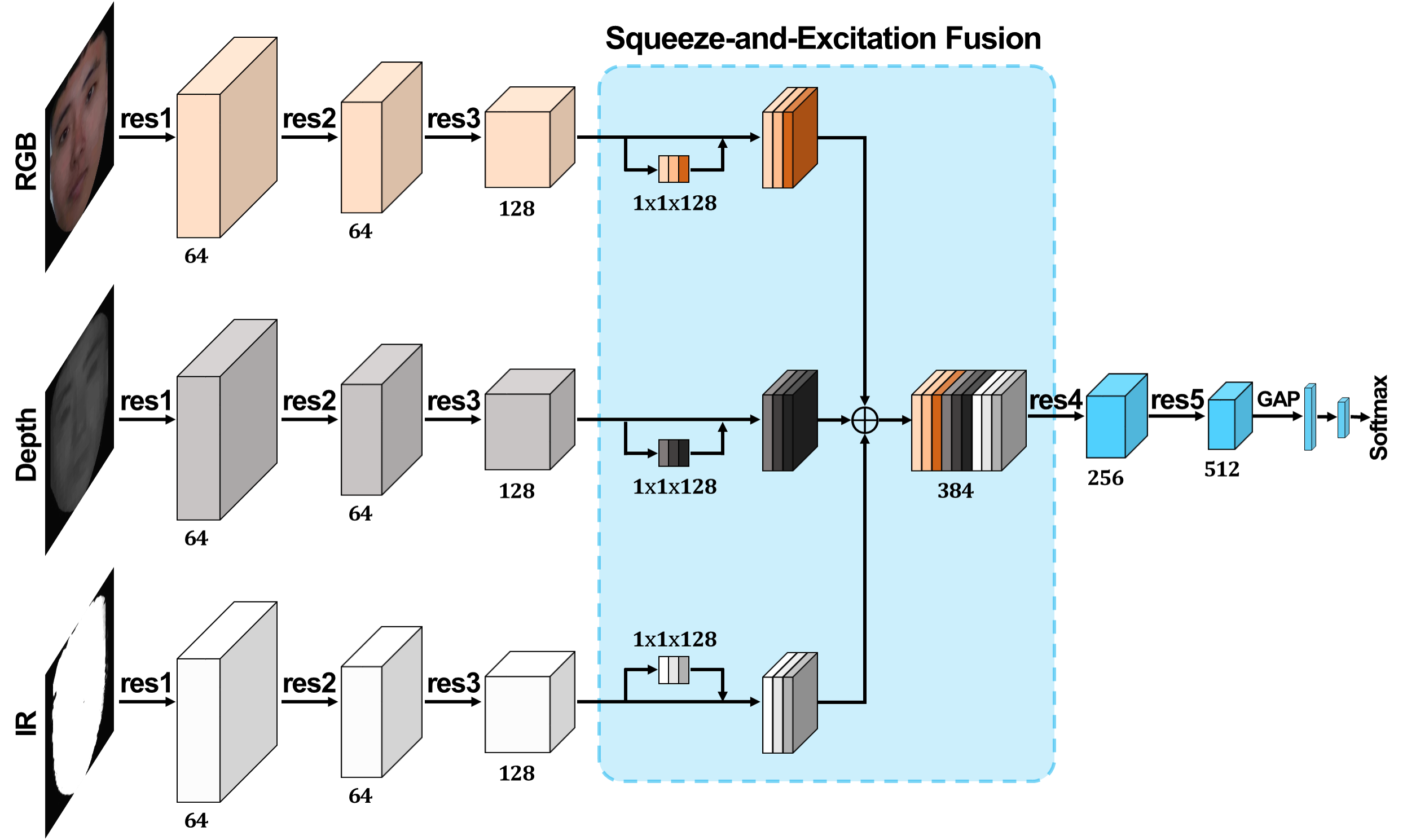}
\caption{Diagram of the proposed fusion method. Each stream uses ResNet-18 as backbone, which has five convolution blocks (\ie, res1, res2, res3, res4, res5). The res1, res2, and res3 blocks extract features of each modal data (\ie, RGB, Depth, IR). Then, these features from different modalities are fused via the squeeze and excitation fusion module. Next, the res4 and res5 block are shared to learn more discriminative features from the fused one. GAP means the global average pooling.}
\label{fig:sef}
\end{figure}

\subsection{Squeeze and excitation fusion}
The three modalities provide with complementary information for different kind of attacks: RGB data have rich appearance details, Depth data are sensitive to the distance between the image plane and the corresponding face, and the IR data measure the amount of heat radiated from a face. Inspired by ~\cite{hu2018senet}, we propose the squeeze and excitation fusion method that uses the ``Squeeze-and-Excitation'' branch to enhance the representational ability of the different modalities' feature by explicitly modelling the interdependencies between their convolutional channels.

As shown in Figure \ref{fig:sef}, our squeeze and excitation fusion method has a three-stream architecture and each subnetwork is feed with the image of different modalities. The res1, res2 and res3 blocks from each stream extract features from different modalities. After that, these features are fused via the squeeze and excitation fusion module. This module newly adds a branch for each modal and the branch is composed of one global average pooling layer and two consecutive fully connected layers. The squeeze and excitation fusion module performs modal-dependent feature re-weighting to select the more informative channel features while suppressing less useful features from each modality. These re-weighted features are concatenated to define the fused multi-modal feature set. 

\begin{table*}
\renewcommand\arraystretch{1.2}
\centering
\setlength{\tabcolsep}{11.4pt}
\small{
\begin{tabular}{|c|c|c|c|c|c|c|}
\hline
\multirow{2}{*}{Method} & \multicolumn{3}{c|}{TPR (\%)} & \multirow{2}{*}{APCER (\%)} & \multirow{2}{*}{NPCER (\%)} & \multirow{2}{*}{ACER (\%)}\\
\cline{2-4}
 & @FPR=$10^{-2}$ &@FPR=$10^{-3}$ &@FPR=$10^{-4}$ & &  & \\
\hline
\hline
Halfway fusion &89.1 &33.6 &17.8 &5.6 &3.8 &4.7 \\
\hline
Proposed fusion &\textbf{96.7} &\textbf{81.8} &\textbf{56.8} &\textbf{3.8} &\textbf{1.0} &\textbf{2.4}\\
\hline
\end{tabular}}
\vspace{1.0mm}
\caption{Effectiveness of the proposed fusion method. All models are trained in the CASIA-SURF training set and tested in the testing set.}
\label{tab:ablation}
\end{table*}

\section{Experiments}
This section describes the implementation details, evaluates the effectiveness of the proposed fusion method, and presents a series of experiments to analyze the CASIA-SURF dataset in terms of modalities and number of subjects. Finally, the generalization capability of a baseline model trained with the CASIA-SURF dataset is evaluated/fine-tuned when tested on standard face anti-spoofing benchmarks.

\subsection{Implementation details}
We resize the cropped face region to $112\times112$, and use random flipping, rotation, resizing, cropping and color distortion for data augmentation. For the CASIA-SURF dataset analyses, all models are trained for $2,000$ iterations with $0.1$ initial learning rate, and decreased by a factor of $10$ after $1,000$ and $1,500$ iterations. All models are optimized via Stochastic Gradient Descent (SGD) algorithm on $2$ TITAN X (Maxwell) GPU with a mini-batch $256$. Weight decay and momentum are set to $0.0005$ and $0.9$, respectively.

\subsection{Model analysis}
We carry out an ablation experiment on the CASIA-SURF dataset to analyze our proposed fusion method. For evaluation, we use the same settings except for the fusion strategy to examine how the proposed method affects final performance. From the results listed in Table~\ref{tab:ablation}, it can be observed that the proposed fusion method achieves TPR=$96.7\%$, $81.8\%$, $56.8\%$ @FPR=$10^{-2}$, $10^{-3}$, $10^{-4}$, respectively, which are $7.6\%$, $48.2\%$ and $39.0\%$ higher than the halfway fusion method, especially at FPR=$10^{-3},10^{-4}$. Besides, the APCER, NPCER and ACER are also improved from $5.6\%$, $3.8\%$ and $4.7\%$ to $3.8\%$, $1.0\%$ and $2.4\%$, respectively. Compared with halfway fusion method, we show the effectiveness of the proposed squeeze and excitation fusion method.

\begin{table*}
\renewcommand\arraystretch{1.2}
\centering
\setlength{\tabcolsep}{11pt}
\small{
\begin{tabular}{|c|c|c|c|c|c|c|c|c|}
\hline
\multirow{2}{*}{Modal} & \multicolumn{3}{c|}{TPR (\%)} & \multirow{2}{*}{APCER (\%)} & \multirow{2}{*}{NPCER (\%)} & \multirow{2}{*}{ACER (\%)}\\
\cline{2-4}
 & @FPR=$10^{-2}$ &@FPR=$10^{-3}$ &@FPR=$10^{-4}$ & &  & \\
\hline
\hline
RGB &49.3 &16.6 &6.8 &8.0 &14.5 &11.3 \\
\hline
Depth &88.3 &27.2 &14.1  &5.1 &4.8 &5.0\\
\hline
IR &65.3 &26.5 &10.9  &15.0 &1.2 &8.1 \\
\hline
RGB\&Depth &86.1 &49.5 &10.6 &4.3 &5.6 &5.0\\
\hline
RGB\&IR &79.1 &50.9 &26.1 &14.4 &1.6 &8.0 \\
\hline
Depth\&IR &89.7 &71.4 &24.3 & \textbf{1.5} &8.4 &4.9\\
\hline
RGB\&Depth\&IR &\textbf{96.7} &\textbf{81.8} &\textbf{56.8} &3.8 &\textbf{1.0} &\textbf{2.4}\\
\hline
\end{tabular}}
\vspace{1.0mm}
\caption{Effect on the number of modalities. All models are trained in the CASIA-SURF training set and tested on the testing set.}
\label{tab:modalities}
\end{table*}

\subsection{Dataset analysis}
The proposed CASIA-SURF dataset has three modalities with $1,000$ subjects. In this subsection, we analyze modalities complementarity when training with a large number of subjects.

{\flushleft \textbf{Effect on the number of modalities. }}
As shown in Table~\ref{tab:modalities}, only using the prevailing RGB data, the results are TPR=$49.3\%$, $16.6\%$, $6.8\%$ @FPR=$10^{-2}$, $10^{-3}$, $10^{-4}$, $8.0\%$ (APCER), $14.5\%$ (NPCER) and $11.3\%$ (ACER), respectively. In contrast, simply using the IR data, the results can be improved to TPR=$65.3\%$, $26.5\%$, $10.9\%$ @FPR=$10^{-2}$, $10^{-3}$, $10^{-4}$, $1.2\%$ (NPCER) and $8.1\%$ (ACER), respectively.
Notably, from the numbers, one can observe that the APCER of the IR data increases by a large margin, from $8.0\%$ to $15.0\%$. Among these three modalities, the Depth data achieves the best performance, \ie, TPR=$88.3\%$, $27.2\%$, $14.1\%$ @FPR=$10^{-2}$, $10^{-3}$, $10^{-4}$, and $5.0\%$ (ACER), respectively. By fusing the data of arbitrary two modalities or all the three ones, we observe an increase in performance. Specifically, the best results are achieved by fusing all the three modalities, improving the best results of single modality from TPR=$88.3\%$, $27.2\%$, $14.1\%$ @FPR=$10^{-2}$, $10^{-3}$, $10^{-4}$, $5.1\%$ (APCER), $1.2\%$ (NPCER) and $5.0\%$ (ACER) to TPR=$96.7\%$, $81.8\%$, $56.8\%$ @FPR=$10^{-2}$, $10^{-3}$, $10^{-4}$, $3.8\%$ (APCER), $1.0\%$ (NPCER) and $2.4\%$ (ACER), respectively. 

\begin{figure}[t]
\centering
\includegraphics[width=0.48\textwidth]{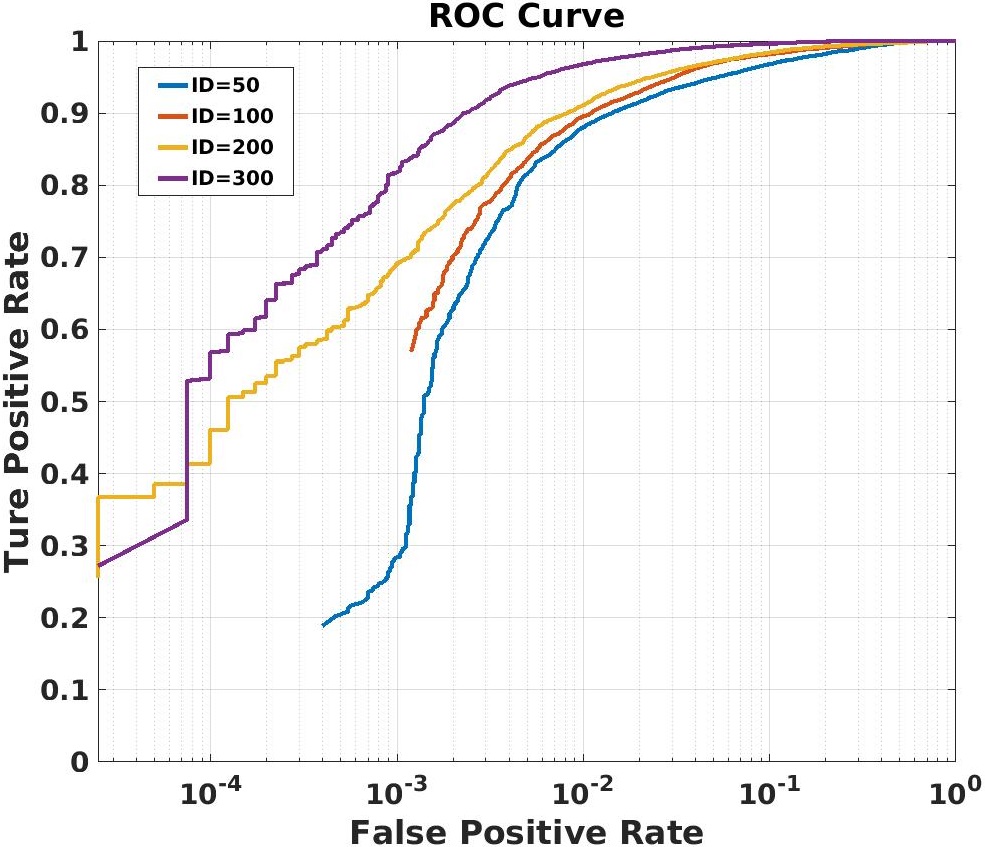}
\caption{ROC curves of different training set size in the CASIA-SURF dataset.}
\label{fig:identity}
\end{figure}

\begin{figure}[t]
\centering
\includegraphics[width=0.47\textwidth]{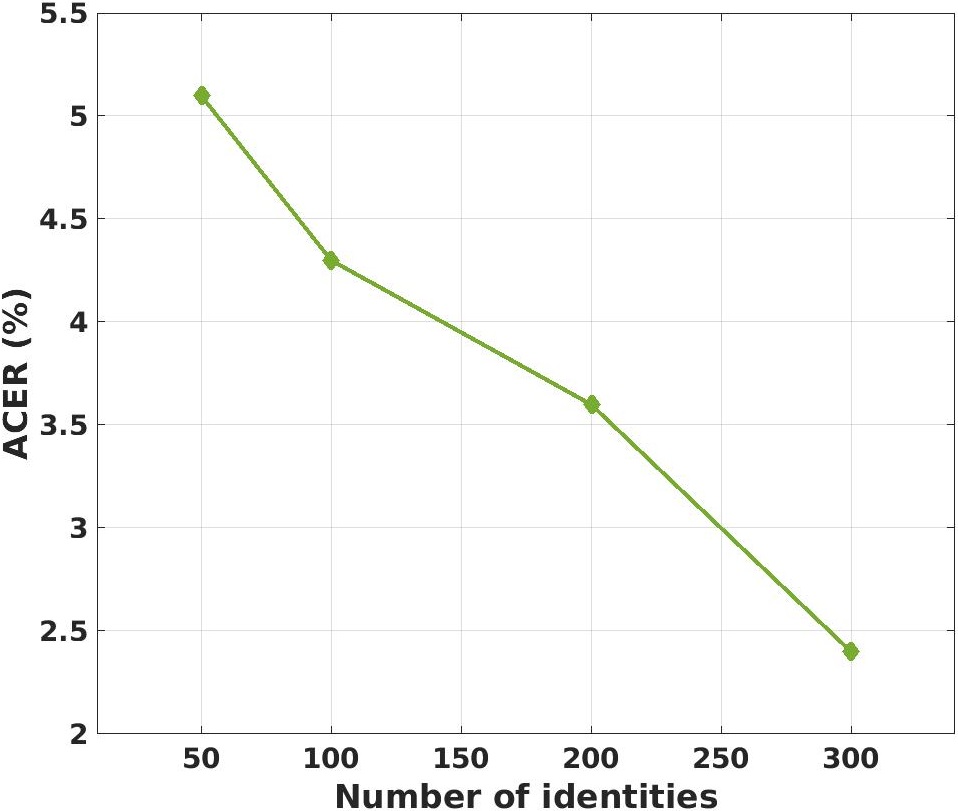}
\caption{Performance \emph{vs.} training set size in the CASIA-SURF dataset.}
\label{fig:acer}
\end{figure}

{\flushleft \textbf{Effect on the number of subjects. }}
As described in~\cite{DBLP:conf/iccv/SunSSG17}, there is a logarithmic relation between the amount of training data and the performance of deep neural network methods. To quantify the impact of having a large amount of training data in PAD, we show how the performance grows as training data increases in our benchmark. For this purpose, we train our baselines with different sized subsets of subjects randomly sampled from the training set. This is, we randomly select $50$, $100$ and $200$ from $300$ subjects for training. Figure\ref{fig:identity} shows the ROC curves for different number of subjects. We can see that when FPR is between $0$ to $10^{-4}$, the TPR is better when more subjects are used for training. Specially, when FPR=$10^{-2}$, the best TPR of 300 subjects is higher about $7\%$ than the second best TPR result (ID=200), showing the more data is used, the better performance will be.
In Figure~\ref{fig:acer}, we also provide with the performance of APCER when a different number of subjects is used for training. The performance of ACER (average value of the fake and real error rates) is getting better when more subjects are considered. 

\subsection{Generalization capability}

In this subsection, we evaluate the generalization capability of a model trained using the proposed dataset when tested/fine-tuned on the SiW~\cite{Liu2018Learning} and CASIA-MFSD~\cite{Zhang2012A} datasets. The CASIA-SURF dataset contains not only RGB images, but also the corresponding Depth information, which is indeed beneficial for Depth supervised face anti-spoofing methods~\cite{Liu2018Learning,FASTD2018arxiv}. Thus, we adopt FAS-TD-SF~\cite{FASTD2018arxiv} as our baseline for the experiments.

{\noindent \textbf{SiW dataset.}} Two state-of-the-art methods (FAS-BAS~\cite{Liu2018Learning} and FAS-TD-SF~\cite{FASTD2018arxiv}) on the SiW dataset are selected for comparison. We use the RGB and Depth images from the proposed CASIA-SURF dataset to pre-train the FAS-TD-SF CNN model, and then fine-tune it in the SiW dataset. Table~\ref{tab:SiW} shows the comparison of these three methods. FAS-TD-SF generally achieves better performance than FAS-BAS, while our pre-trained FAS-TD-SF in CASIA-SURF (FAS-TD-SF-CASIA-SURF) can further improve the performance of PAD on both protocols\footnote{For more details of the protocols, please refer to \cite{Liu2018Learning}.} 1, 2 and 3. Concretely, the performance of ACER is superior about $0.25\%$, $0.14\%$ and $1.38\%$ when using the proposed CASIA-SURF dataset in Protocol 1, 2, and 3, respectively. The improvement indicates that pre-training in the CASIA-SURF dataset supports the generalization on data containing variabilities in terms of (1) face pose and expression, (2) replay attack mediums, and (3) cross presentation attack instruments (PAIs), such as from print attack to replay attack. Interestingly, it also demonstrates our dataset is also useful to be used for pre-trained models when replay attack mediums cross PAIs. 

\begin{table}[h]
\renewcommand\arraystretch{1.15}
\centering
\setlength{\tabcolsep}{2.2pt}
\footnotesize{
\begin{tabular}{|c|c|c|c|c|}
\hline
Prot. & Method & APCER(\%) & BPCER(\%) & ACER(\%) \\
\hline
\hline
\multirow{3}{*}{1} &FAS-BAS~\cite{Liu2018Learning} &3.58 &3.58 &3.58 \\
\cline{2-5} &FAS-TD-SF~\cite{FASTD2018arxiv} &1.27 &0.83 &1.05 \\
\cline{2-5} &\textbf{FAS-TD-SF-CASIA-SURF} &1.27 &0.33 &\textbf{0.80} \\
\hline
\multirow{3}{*}{2} &FAS-BAS~\cite{Liu2018Learning} &0.57$\pm$0.69 &0.57$\pm$0.69 &0.57$\pm$0.69 \\
\cline{2-5} &FAS-TD-SF~\cite{FASTD2018arxiv} &0.33$\pm$0.27 &0.29$\pm$0.39 &0.31$\pm$0.28 \\
\cline{2-5} &\textbf{FAS-TD-SF-CASIA-SURF} &0.08$\pm$0.17 &0.25$\pm$0.22 &\textbf{0.17$\pm$0.16} \\
\hline
\multirow{3}{*}{3} &FAS-BAS~\cite{Liu2018Learning} &8.31$\pm$3.81 &8.31$\pm$3.80 &8.31$\pm$3.81 \\
\cline{2-5} &FAS-TD-SF~\cite{FASTD2018arxiv} &7.70$\pm$3.88 &7.76$\pm$4.09 &7.73$\pm$3.99 \\
\cline{2-5} &\textbf{FAS-TD-SF-CASIA-SURF} &6.27$\pm$4.36 &6.43$\pm$4.42 &\textbf{6.35$\pm$4.39} \\
\hline
\end{tabular}
}
\vspace{-1mm}
\caption{Evaluation results in three protocols of SiW. }
\label{tab:SiW}
\end{table}

{\noindent \textbf{CASIA-MFSD dataset.}} Here, we perform cross-testing experiments on the CASIA-MFSD dataset to further evaluate the generalization capability of models trained with the proposed dataset. State-of-the-art models~\cite{Pereira2013Can, Bharadwaj2013Computationally, pinto2015face, yang2014learn} used for comparison used Replay-Attack~\cite{Chingovska_BIOSIG-2012} for training. We then train the FAS-TD-SF~\cite{FASTD2018arxiv} in the SiW and CASIA-SURF datasets. Results in Table~\ref{tab:cross-testing} show that the model trained in the CASIA-SURF dataset performs the best. 

\begin{table}[h]
\renewcommand\arraystretch{1.15}
\centering
\setlength{\tabcolsep}{5pt}
\footnotesize{
\begin{tabular}{|c|c|c|c|}
\hline
Method & Training & Testing & HTER (\%)\\
\hline
\hline
Motion~\cite{Pereira2013Can} &Repaly-Attack &CASIA-MFSD &47.9 \\
\hline
LBP~\cite{Pereira2013Can} &Repaly-Attack &CASIA-MFSD &57.6 \\
\hline
Motion-Mag~\cite{Bharadwaj2013Computationally} &Repaly-Attack &CASIA-MFSD &47.0 \\
\hline
Spectral cubes~\cite{pinto2015face} &Repaly-Attack &CASIA-MFSD &50.0\\
\hline
CNN~\cite{yang2014learn} &Repaly-Attack &CASIA-MFSD &45.5\\
\hline
FAS-TD-SF~\cite{FASTD2018arxiv} &SiW &CASIA-MFSD &39.4\\
\hline
\textbf{FAS-TD-SF}~\cite{FASTD2018arxiv} &\textbf{CASIA-SURF} &\textbf{CASIA-MFSD} &\textbf{37.3}\\
\hline
\end{tabular}
}
\vspace{0.05mm}
\caption{Cross testing results on different cross-testing protocols.}
\label{tab:cross-testing}
\end{table}

\section{Discussion}
As shown in Table~\ref{tab:ablation} and Table~\ref{tab:modalities}, accurate results were achieved in the CASIA-SURF dataset for traditional metrics, e.g. APCER=$3.8\%$, NPCER=$1.0\%$, ACER=$2.4\%$. However, this shows an error rate of fake samples of $3.8\%$ and an error rate of real samples of $1.0\%$. Thus, 3.8 fake samples from 100 attackers will be treated as real ones. This is below the accuracy requirements of real applications, e.g., face payment and phone unlock. Table~\ref{tab:SiW} also demonstrates a similar performance in the SiW dataset. In order to push the state-of-the-art, in addition to large datasets, new evaluation metrics would be beneficial. The ROC curve is widely used in academic and industry for face recognition~\cite{liu2017sphereface}. We consider the ROC curve to be also appropriated to be used as evaluation metric for face anti-spoofing.

As shown in Table~\ref{tab:ablation} and Table~\ref{tab:modalities}, although the value of ACER is very promising, the TPR at different values of FPR is dramatically changing, being far from the standard required in real applications, e.g. when FPR=$10^{-4}$ the TPR is $56.8\%$. Similar to the evaluation of face recognition algorithms, the TPR when FPR is about $10^{-4}$ or $10^{-5}$ would be meaningful for face anti-spoofing~\cite{kemelmacher2016megaface}. 

\section{Conclusion}
In this paper, we presented and released a large-scale multi-modal face anti-spoofing dataset. The CASIA-SURF dataset is the largest one in terms of number of subjects, data samples, and number of visual data modalities. We believe this dataset will push the state-of-the-art in face anti-spoofing. Owing to the large-scale learning, we found that traditional evaluation metrics in face anti-spoofing (\ie, APCER, NPECR and ACER) did not clearly reflect the utility of models in real application scenarios. In this regard, we proposed the usage of the ROC curve as the evaluation metric for large-scale face anti-spoofing evaluation. Furthermore, we proposed a multi-modal fusion method, which performs modal-dependent feature re-weighting to select the more informative channel features while suppressing the less informative ones. Extensive experiments have been conducted on the CASIA-SURF dataset, showing high generalization capability of models trained on the proposed dataset and the benefit of using multiple visual modalities.

\section{Acknowledgements}
This work has been partially supported by the Science and Technology Development Fund of Macau (Grant No. 0025/2018/A1), by the Chinese National Natural Science Foundation Projects $\#$61876179, $\#$61872367, by JDGrapevine Plan in the JD AI Research, by the Spanish project TIN2016-74946-P (MINECO/FEDER, UE) and CERCA Programme / Generalitat de Catalunya, and by ICREA under the ICREA Academia programme. We gratefully acknowledge Surfing Technology Beijing co., Ltd (www.surfing.ai) to capture and provide us this high quality dataset for this research. We also acknowledge the support of NVIDIA with the GPU donation for this research.

{\small
\bibliographystyle{ieee_fullname}
\bibliography{egbib}
}

\end{document}